# TEAM: An Taylor Expansion-Based Method for Generating Adversarial Examples

QIAN Ya-guan[1], ZHANG Xi-Ming[1], Wassim Swaileh[5], Li Wei[1], WANG Bin[2], CHEN Jian-Hai[3], ZHOU Wu-Jie[4], LEI Jing-Sheng[4]

[1](*School of Big-Data Science, Zhejiang University of Science and Technology, Hangzhou 310023, China*)

[2](*Network and Information Security Laboratory of Hangzhou Hikvision Digital Technology Co., Ltd. Hang Zhou 310052, China*)

[3]*(College of Computer Science and Technology, Zhejiang University, Hangzhou 310058, China)*

[4](*School of Information and Electronic Engineering, Zhejiang University of Science and Technology, Hangzhou 310023, China*)

[5]*(Laboratoire ETIS UMR-CNRS 8051, CY Cergy Paris Université, France)*

∗Corresponding author
Email address: qianyaguan@zust.edu.cn (QIAN Ya-guan)

**Abstract:** Although Deep Neural Networks(DNNs) have achieved successful applications in many fields, they are vulnerable to adversarial examples.Adversarial training is one of the most effective methods to improve the robustness of DNNs, and it is generally considered as solving a saddle point problem that minimizes risk and maximizes perturbation.Therefore, powerful adversarial examples can effectively replicate the situation of perturbation maximization to solve the saddle point problem.The method proposed in this paper approximates the output of DNNs in the input neighborhood by using the Taylor expansion, and then optimizes it by using the Lagrange multiplier method to generate adversarial examples. If it is used for adversarial training, the DNNs can be effectively regularized and the defects of the model can be improved.

## 1 Introduction

At present, deep neural networks (DNNs) are applied successfully in many fields such as bioinformatics [1,2], speech recognition [3,4], and computer vision [5,6]. However, recent work has shown that DNNs are vulnerable to adversarial examples. Szeigy et al. [7] first observed adversarial examples of DNNs in the image classification domain. An adversarial example is a natural image maliciously added by tiny perturbations that is almost imperceptible to human eyes, which can fool the DNNs to produce an incorrect prediction. An typical example is in the autonomous driving domain, where a traffic signs crafted as an adversarial example, can successfully deceive the DNN of autonomous car, which can cause serious traffic accidents.

Adversarial examples attract a lot of interest of researchers and many typical attack methods are proposed to generate them such as FGSM [8], JSMA [9], Deepfool[10], C&W [11], PGD [12], M-DI$^2$-FGSM[13] etc. Meanwhile, some defense methods are also suggested. Adversarial training is first proposed by Szeigy et al.[15] , which is one of the most effective defense proved by [z1] so far. Ruitong Huang et al. [12,15,16] from optimization viewpoint, consider adversarial training is a min-max optimization problem. The key to address this problem is to generate more powerful adversarial examples to maximize the adversarial loss of DNNs. Based on this idea, we propose a novel approach termed as TEAM (Taylor Expansion-based Adversarial Method) that are different from previous proposed methods.

The generation of adversarial examples is usually be modeled as an optimization problem of the loss function with respect to the input. However, it is very difficult to solve this optimization problem due to the high nonlinearity of DNNs. Therefore, the current typical gradient-based methods, such as FGSM, JSMA, Deepfool, C&W, PGD, M-DI$^2$-FGSM etc., all transform it into an easily optimized objective function, which generates the adversarial examples through one or more steps of iteration[9,11]. The objective functions of these methods all depend on the first-order gradient information of DNNs. As a result, FGSM and Deepfool are prone to fall into the local optimal value. JSMA recalculates the gradient every iteration in the last hidden layer. C&W also iterates on the selection of the super parameter. PGD will lose the learned information during the iteration.

Our proposed TEAM is to leverage a quadratic function to replace the nonlinear part of DNNs in a small neighbourd of the input example by Taylor expansion. Since it approximates the input-output mapping relationship of DNNs in a tiny neighbourhood of the input, the problem of generation of DNNs' adversarial examples is translate to the problem of generation of quadratic functions' adversarial examples, which is equivalent from the view of classification. After that, we further thanslate it to a dual problem on the basis of Lagrange multiplier method. At this point, we turn the original problem into a convex problem which is more easy to resolve.

Compared with the one-step attack method like FGSM, our method is closer to the optimal solution, which makes adversarial examples more powerful. Compared with other multi-step attack methods, our method has advantages in calculation accuracy, and can avoid falling into local optimization. Therefore, our method can not only generate powerful adversarial examples, but also significantly reduce the computing complexity.

Transferability is an important property of adversarial examples, which is used by adversary to launch a black-box attack. That is, the adversarial examples generated by one DNNs is equally effective on another DNNs. Florian et al.[17] have deeply explored the transferability of adversarial examples. Xie et al.[13] comprehensively studied and analyzed the transferability of adversarial examples generated by different attack methods. Hence, in our experiments, besides attack success rates, transferability is another measure to evaluate our method compared with other methods. In experimental results, the attack success rates of our method could achieve 100% both on MNIST and CIFAR10 data sets. In addition, transferability of our method could achieve 68% and 65% on MNIST and CIFAR10 data sets respectively, which are higher than other methods.

Equipped with above perspective, we make the following contributions:
(1) Through analysis, we found that adversarial examples need to keep good concealment, so effective adversarial examples must be obtained in a small perturbation range. For this reason, we only need to apply Taylor theorem in the small field ($L_p$ norms constraints) of the input example, generating a quadratic Taylor expansion which has similar output to DNNs. Then, the Lagrange multiplier method is used to construct the dual function for optimization. Compared with all the previous methods, this method can avoid being trapped by local optimizations, and use dual function to optimize can get more effective adversarial examples.

(2) We systematically evaluated the selection of the objective function to construct adversarial examples, and proved through experiments that our method can generate effective adversarial examples whether DNNs smooth the gradient for defense or not.

(3) We propose using our high-confidence adversarial examples in adversarial training test, and prove that our method overperforms the most representative attack methods available nowadays in the aspect of. improving the robustness of DNNs and the transferability adversarial examples

Experimental evaluation was conducted on MNIST and CIFAR10 data seta. To make it easier for its researchers to use our work to evaluate the robustness of other defense systems, The complete code is available at https://github.com/zhangximin2019/zhangximin.

This paper is organized as follows: Sec. 2 introduces background on the fundamentals of DNNs and the most representative attack methods available nowadays. Sec. 3 describes TEBM and another method based on the Gauss-Newton Method in detail, providing mathematical derivation and algorithm. Sec. 4 discusses the inner maximization problem and outer minimization problem in adversarial training. Sec.5 carries on the experiment evalueation, with experimental analysis to verify the effectiveness of our method. In Sec. 6 briefly describes the related work of this paper. In Sec. 7 we conclude our work.

## 2  Background

This section provides background on our work, covering the fundamentals of DNNs and the most representative attack methods available nowadays.

### 2.1  Deep Neural Networks and Notation

DNNs can generally be represented as a multidimensional function $F: \mathcal{X} \mapsto \mathcal{Y}$, $X \in \mathcal{X}$ stands for $d$-dimensional input variable, $Y \in \mathcal{Y}$ is a $h$-dimensional probability vector which stands for Confidence of $h$ classes. An $N$-layer DNNs receives an input $X$ and produces the corresponding output as follows:

$$F(X) = F^{(N)}(\ldots F^{(2)}(F^{(1)}(X))) \tag{1}$$

$F^{(i)}$ represents the computational output of layer $i$ of DNNs. These layers can be convolutional, pooled, or other forms of neural network layers. The last layer of DNNs is usually the Softmax layer, defined as $F^{(N)}(Z)_i = \text{Softmax}(Z)_i = \exp(z_i) / \sum_{i=1}^{m} \exp(z_i)$, $Z = F^{(N-1)}(\cdot)$ is the output vector of the previous layer (also known as the last hidden layer). The final prediction label is obtained by $y = \arg\max_{i=1\ldots m} F(X)_i$, where $F(X) = \text{Softmax}(Z)$.

## 2.2 Adversarial Examples of DNNs

Szegedy et al. [7] first found the existence of adversarial examples in DNNs. More formally, in the space $\mathbb{R}^{m \times n}$, we think of an image $X$ which size is $m \times n$ as a point. Our goal is to find a point $X' = X + \delta, \|\delta\|_p < C$ which is in the constraint and is close enough to $X$. Such point $X'$ is called the adversarial example. This $X'$ and $X$ belong to the same category from the perspective of human eye, but the subtle perturbation $\delta$ deceives DNNs into judging it as a different class from $X$, i.e

$$F(X') = F(X + \delta) = y' \quad s.t. \quad y' \neq y \tag{2}$$

In general, the perturbation $\delta$ is constrained by the $L_p$ norm ($p \in 0, 2, \infty$), $\|X' - X\|_p \leq C$.

## 2.3 Threat Models in Deep Learning

There are a number of methods to generate adversarial examples, but they are all have constraints. Since the capability of adversarial examples or the robustness of the attack is based upon what adversary is allowed to do. Without such limitation, adversary can replace the given image with any image, violating the definition of adversarial examples. To this end, we define these assumptions as threat models, which typically include attack targets and attack capabilities.

(1) Adversarial Goals

Adversarial goals in threat models can be defined as a specific formula that needs to be detected and defended. In DNNs, the classification of adversarial Goals is helpful for us to define this specific formula. Therefore, in threat models, the classification of adversarial goals is very important. In this paper, adversarial goals are divided into two categories:

a) Untargeted attack: misclassify adversarial examples to any incorrect class
b) Targeted attack: misclassify adversarial examples to specified incorrect class

In this paper, both untargeted attack and targeted attack are based on the change of confidence, and the method in this paper is the first attack method to reduce confidence through optimization method from the mathematical perspective to generate adversarial examples.

(2) Adversarial Capabilities

Adversarial examples can also be divided into white box attack and black box attack according to how much information adversary has about the target DNNs. The so-called white box attack means that adversary knows everything about DNNs, including training data, activation function, topology structure, weight coefficient and so on. The black box attack assumes that adversary cannot obtain the internal information of the target DNNs and can only obtain the output of the model, including labels and confidence.

Because the gradient information of target DNNs needs to be mastered, the method in this paper belongs to white box attack. However, since adversarial examples generated by the method in this paper is highly transferable, it is easy to build the agent DNNs locally through the method in the paper [18] to successfully realize the black box attack.

## 2.4 Attack algorithms

We selected some typical gradient-based methods to compare with the methods proposed in this paper. The existing typical methods include FGSM, JSMA, Deepfool, C&W, PGD, M-DI$^2$-FGSM, etc.

**(1) FGSM**

Goodfellow et al.[8] presents a method for rapidly generating adversarial examples under $L_\infty$ distance called FGSM(Fast Gradient Sign Method):

$$X' = X + \varepsilon \cdot \text{sign}(\nabla_X J(F(X;\theta), y)) \tag{3}$$

where $J$ is the loss function, $\varepsilon$ is the perturbation limit on the sign gradient direction $\text{sign}(\cdot)$. FGSM has the advantages of low computational complexity and the ability to generate a large number of adversarial examples in a short time.

**(2) JSMA**

Papernot et al.[9] proposed a targeted attack method under $L_0$ distance. According to the adversarial saliency maps, the input components were ranked in descending order, and the components with strong adversarial saliency were selected to add perturbation $\delta$. For the target class $t$, adversarial saliency map of component $S(X,t)[i]$ is defined as:

$$S(X,t)[i] = \begin{cases} 0, & \text{if } \frac{\partial F_t(X)}{\partial x_i} < 0 \text{ or } \sum_{j \neq t} \frac{\partial F_j(X)}{\partial x_i} > 0 \\ (\frac{\partial F_t(X)}{\partial x_i}) \left| \sum_{j \neq t} \frac{\partial F_j(X)}{\partial x_i} \right|, & \text{otherwise} \end{cases} \tag{4}$$

where $J_F = \left[ \partial F_j(X) / \partial x_i \right]_{ij}$ represents DNNs Jacobian matrix. During each iteration, the component $X_i$ of the maximum value of the adversarial saliency value is selected to increase by a constant offset until the example is misclassified.

**(3) Deepfool**

Moosav-dezfooli et al.[10] proposed the Deepfool method, which employs linear approximation for gradient iterative attack. For a binary classifier, the following iterative process can be used to describe:

$$\begin{aligned} \delta_i &= -\frac{f(X'_i)}{\|\nabla f(X'_i)\|_2^2} \nabla f(X'_i) \\ X'_{i+1} &= X'_i + \delta_i \\ \delta &= \sum_i \delta_i \end{aligned} \tag{5}$$

Here $X'_i$ is the adversarial examples for the $i$-th iteration, $X'_0 = X$. $f(X'_i) / \|\nabla f(X'_i)\|_2$ is the estimated distance between $X'_i$ and the decision boundary $f(X'_i)$. $-f(X'_i)/\|\nabla f(X'_i)\|_2$ is the gradient direction of $X'_i$ toward the decision boundary.

**(4) C&W**

Carlini and Wagner proposed a targeted iterative attack method based on gradient descent. Based on their further studies[11,19,20], C&W attacks are effective against most existing defenses. They modeled the process of generating the adversarial examples as the following optimization problem, minimizing the disturbance while maximizing the model classification error:

$$\min_{\delta} \|\delta\|_p + c \cdot g(X+\delta) \qquad (6)$$

If and only if $F(X') = y'$, $g(X') \geq 0$. Through experimental evaluation, they found that the most effective function $g$ was:

$$g(X') = \max(\max_{i \neq y'}(\text{Softmax}(X')_i) - \text{Softmax}(X')_t, -k) \qquad (7)$$

Where $k$ is the constant that controls confidence.

**(5) PGD**

Aleksander et al.[12] explained the process of generating adversarial examples as a simple one-step solution to solve the internal maximization problem of saddle point problem. Based on this, they proposed a derivative method of FGSM, called PGD. The essence of this method is to project gradient descent on the loss function:

$$X' = \prod_{X+S}(X + \varepsilon \cdot \text{sign}(\nabla_X J(f(X;\theta), y))) \qquad (8)$$

**(6) M-DI$^2$-FGSM**

Momentum and diverse inputs are two completely different ways to alleviate the overfitting phenomenon. Xie et al.[13] combined them naturally to form a much stronger attack, i.e., Momentum Diverse Inputs Iterative Fast Gradient Sign Method (M-DI$^2$-FGSM):

$$g_{n+1} = \mu \cdot g_n + \frac{\nabla_X L(T(X_n^{adv}; p), y^{true}; \theta)}{\|\nabla_X L(T(X_n^{adv}; p), y^{true}; \theta)\|_1} \qquad (9)$$

$$X'_{n+1} = \text{Clip}_X^{\varepsilon}\{X'_n + \alpha \cdot \text{sign}(\nabla_X L(X'_n, y^{true}; \theta))\} \qquad (10)$$

## 3  Our approach

We present algorithm algorithm1 to find appropriate perturbation $\delta$ under $L_p$ condition. For the untargeted attack, the algorithm minimizes the confidence of classifying an AEs towards its correct class. However, it maximizes the classification confidence toward a target class in the targeted attack.

To achieve this goal, we conducted quadratic Taylor Expansion to approximate the complicated input-output mapping relationship of DNNs in a small neighbourhood of input example. Then, we used the Lagrange multiplier method to construct the dual function that produces the perturbation $\delta$.

For untargeted attack, it is to reduce the confidence of the correct output class; for targeted attack, it is to improve the confidence of the target class. Therefore, the core of the proposed method is constructing a quadratic Taylor expansion to approximate the complicated input output projection of DNNs in a small neighbourhood of the input example by using gradient information. Then, the Lagrange multiplier method is used to construct the dual function and calculate the extreme value to obtain the perturbation $\delta$.

First, our method calculates the gradient information of DNNs by considering that the output of the last hidden layer (before and after Softmax layer) of DNNs can both provide gradient information of DNNs. In contrast, the output after the Softmax layer is the normalisation of the output before the softmax. The Softmax layer smooth the gradient. The author of [9] insist on using the last hidden layer output (before the softmax) for the calculation. Indeed, the extreme variations introduced by the logistic regression computed between these two layers leads to absolute derivative values. This reduces the quality of information on how different inputs activate the neurons according to[9]. For evaluating our method, We extracted "Logist vector" from the outputs of the last hidden layer (before the softmax layer), and "Confidence vector" from the Softmax output.

By considering that we obtain AEs using a small neighbourhood of the input. In this neighbourhood, DNNs can be easily trapped by local optimization. That because of it still highly nonlinear even if we used multi-step iterations methods. To avoid this problem, we used quadratic Taylor Expansion to approximate the complicated input-output mapping relationship of DNNs in a small neighbourhood of the input. A simple quadratic function replaces the nonconvex and nonlinear part of DNNs. Then the Lagrange multiplier method is used to construct the dual function and transform the original problem into a convex problem. In this way, we obtained the effective perturbation value to generate hight attack AEs.

## 3.1 Problem Formulation

In this section, we define the objective functions and the optimisation model for the targeted and untargeted attacks. To obtain AEs from the last hidden layer of DNNs, we considered the output of each neuron of the last hidden layer. Formally, this output is the logist value of $z_i$ that assigned to the class $m$ at the neuron $i$.

For a given input $X$, The highest value of $z_i$ decides the class label $m$ that corresponds to $X$ according to eq.11:

$$m = \underset{i}{\operatorname{argmax}}(z_i(X)) \tag{11}$$

This section will choose the objective function and establish the optimization model according to the attack target in 2.2. Assume that adversarial examples are generated on the last hidden layer. The output of each neuron in the last hidden layer of DNNs is the logist value assigned to the class that the neuron represents, and the predict label of $X$ is $y = \arg\max_{i=1,\ldots,m}(z_i)$. That is, the greater the value of $z_i$, the more likely it is to be classified as the $i$-th class, and vice versa. Since the purpose of adversarial examples, $X'$ is to add a small amount of distortion $\delta$ that fool the DNNs. Thus, our goal is to find $\delta$ that ensure smaller value for $z_m(X+\delta)$. In other words, the low probability of affecting the correct class label $m$ to the input increases the probability of DNNs misclassification.

Alternatively, we can generate AEs using the normalised last DNNs layer outputs (softmax output). Let us denote the softmax output as $f_j(X)$ to distinguish it from the pure last DNNs layer output $z_j(X)$. The difference between the outputs before and after applying the Softmax is that the drop in $f_j(X)$ corresponds to the rise in $\sum_{i \neq j} f_i(X)$. Intuitively, reducing the confidence of the correct class at the Softmax layer is recommended to generate AEs. Our results show no significant difference between using the pure versus normalised last DNNs layer outputs. We can say that our method is more adaptable than the first-order gradient information based attack represented by JSMA.

We considered the problem of obtaining the optimal perturbation $\delta$ as an optimisation problem. For that, we defined three different kinds of objective functions:

I. $T_1 = z_m(X+\delta)$ and $T_4 = f_m(X+\delta)$ represent the output of the correct class of $X+\delta$ at the last hidden layer before and after applying the Softmax, respectively. We can obtain an AEs, once a minimum value of $T_1$ or $T_4$ is obtained.

II. $T_2 = z_t(X+\delta)$ and $T_5 = f_t(X+\delta)$ represent the output of the target class of $X+\delta$ before and after the Softmax. We can obtain an AEs once $T_2$ or $T_5$ reachs its maximum.

III. $T_3 = z_t(X+\delta) - \max_{i \neq t}(z_i(X+\delta))$ and $T_6 = f_t(X+\delta) - \max_{i \neq t}(f_i(X+\delta))$ represent the difference between the target class output value and the maximum class output value at the last hidden layer before and after applying the Softmax. Once $T_3$ and $T_6$ reache their maximum value, the probability of generating targeted adversarial example is maximum.

（1） Untargeted Atteck

IV. Our goal is to find $\delta$ that minimizes $f_m(X+\delta)$ that keep the AEs untractable. Formaly, we are looking for finding the perturbation value $\delta$ for the image $X$, we can write:

$$\begin{aligned}&\text{minimize } T_1 \text{ or } T_4\\&\text{s.t.} \quad F(X+\delta)\neq m,\ X+\delta\in[0,1]^n,\ \|\delta\|_p<C\end{aligned} \quad (12)$$

(2) Targeted Attack

Suppose $t$ is the target class, then the goal is to find the $\delta$ that maximizes $f_t(X+\delta)$ within the constraint:

$$\begin{aligned}&\text{maxmize } T_2 \text{ or } T_3 \text{ or } T_5 \text{ or } T_6\\&\text{s.t.} \quad F(X+\delta)=t,\ X+\delta\in[0,1]^n,\ \|\delta\|_p<C\end{aligned} \quad (13)$$

This paper studies a special input case, that is, the benign sample $X$ is not a meaningful natural picture, it may be a pure black or white picture, or it may be a meaningless messy code. Our adversarial examples can trick DNNs and classify adversarial examples into our target class $t$. We add a condition $F(X)\neq i$ to equation (13).

### 3.2  Generate adversarial examples based on the Taylor expansion

We propose a novel adversarial example generation algorithm1, that proved its effectiveness through experiments. Remember that both objective function $z_i$ and $f_i$ presented in section 3.1 refers to the extracted gradient information. Therefore, for convenience of narration, we denote $f_i$ and $f_i$ uniformly by $F$ in this section.

We use quadratic Taylor Expansion to approximate the nonlinear part of $F$ in the neighbourhood of $X$. It is by transforming the constrained nonlinear optimisation problem into a constrained linear optimisation problem. After that, we use Lagrange multiplier method to construct a dual problem that finds the optimal solution within the $L_p$ norms constraints. This process not only reduces the difficulty of solving the optimisation problem, but it also improves the solution accuracy.

(1) Compute the gradient matrix $\nabla F(X)$ and Hessian matrix $\nabla^2 F(X)$ for the given sample X.

$$\nabla F_m(X)=\left[\frac{\partial F_m(X)}{\partial X_i}\right]_{n\times 1} \quad (14)$$

$$\nabla^2 F_m(X)=\left[\frac{\partial^2 F_m(X)}{\partial X_i \partial X_j}\right]_{n\times n} \quad (15)$$

(2) Use Taylor expansion to approximate $F_m$ in the neighbourhood $U(X,\delta)$ of $X$.

$$F_m(X')=F_m(X+\delta)\approx T(\delta)=F_m(X)+\nabla F_m(X)^T\delta+\frac{1}{2}\delta^T\nabla^2 F_m(X)\delta \quad (16)$$

$X'$ is the moving point in the neighbourhood. $F_m(X')$ is the logist or Softmax value of $X'$

that is classified as the $m$-th class. The minimum value of logist/softmax being, maximise the probability of misclassifying the input sample.

(3) Calculate $\delta$ by using the Lagrange multiplier method: In equation (16), $\delta$ is the only unknown. Therefore, we transform the problem of generating adversarial examples into a nonlinear optimization problem under inequality constraints:

$$\min T(\delta) \quad s.t. \quad \|\delta\|_p^2 \leq C \tag{17}$$

We construct the Lagrange function:

$$\begin{aligned} L(\delta, \lambda) &= T(\delta) + \lambda(\|\delta\|_p^2 - C) \\ &= F_m(X) + \nabla F_m(X)^T \delta + \frac{1}{2}\delta^T \nabla^2 F_m(X)\delta + \lambda(\|\delta\|_p^2 - C) \end{aligned} \tag{18}$$

We transformed the nonlinear optimisation problem with inequality constraints into an unconstrained optimisation problem. To simplify the calculations, we turned the original problem (17) into a dual problem:

$$\min -g(\lambda) \quad s.t. \quad \lambda \geq 0 \tag{19}$$

Where the dual function $g(\lambda) = \inf L(\delta, \lambda)$. According to the weak duality property, the optimal value $d^*$ of equation (19) is the optimal lower bound of the original problem (17), that is, the convex optimization problem approximates the original problem. The optimal solution of equation (19) must satisfy the KKT condition as follows:

$$\begin{cases} \nabla_\lambda - g(\lambda) = 0 \\ \lambda \geq 0 \end{cases} \tag{20}$$

Assume that when $\lambda = \lambda^*$, $-g(\lambda^*)$ can get the minimum; when $\delta = \delta^*$, $T(\delta^*)$ be able to get the minimum. According to the principle of weak duality property, the optimal value of the original problem is not less than the optimal value of the dual problem, i.e. $g(\lambda^*) \leq T(\delta^*)$. If the original function is a convex function and satisfies the Slater condition, then $g(\lambda^*) = T(\delta^*)$. However, due to the high nonlinearity of DNNs, $\nabla^2 F(X)$ is difficult to be proved as a positive definite matrix. So we can think of $(\delta^*, \lambda^*)$ as an approximately optimal solution to $L(\delta, \lambda)$. Figure 1 shows the relationship between DNNs and dual functions.

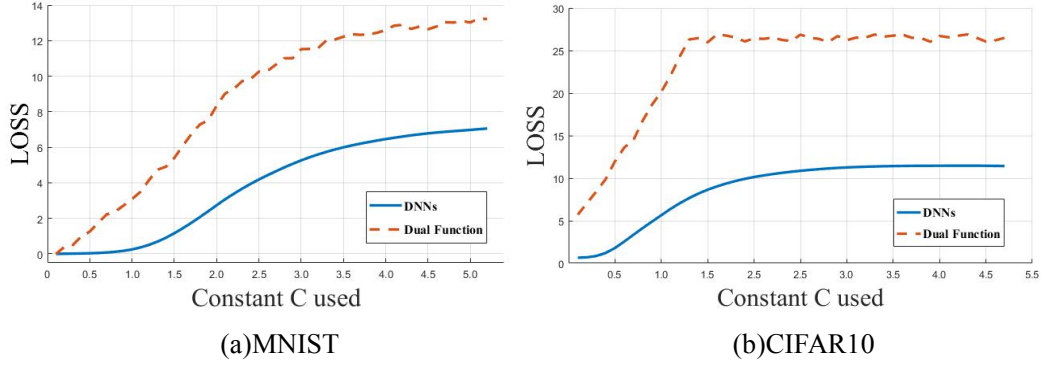

(a)MNIST          (b)CIFAR10

Figure 1: The relationship between DNNs and dual functions. This picture plots the process of generating adversarial example on MNIST and CIFAR10 dataset. For the target function $T_1$, with the constant $C$ increasing, the loss value of DNNs and the dual function changes. We can see that the changes of DNNs and dual functions tend to be consistent.

If the obtained $\delta$ satisfies the condition $\exists t \in I, F_t(X+\delta) > F_y(X+\delta)$, then $\delta$ is the best perturbation to generate adversarial example. In this way, we obtain the optimal solution $\delta$ for the optimization problem with inequality constraints through the Lagrange multiplier method, thus generating the adversarial example $X'$. The above method can be easily extended to all non-cyclic DNNs. The only requirement is that the activation function is differentiable, which has been satisfied by the characteristics of BP algorithm. The whole process is shown in Algorithm 1.

**Algorithm1** Generate adversarial examples based on Taylor expansion

$X$ is a benign example

**Input:** $X, C$

**Output:** $X'$

1: $\nabla F_m(X) \leftarrow \left[\dfrac{\partial F_m(X)}{\partial X_i}\right]_{n\times 1}, \nabla^2 F_m(X) \leftarrow \left[\dfrac{\partial^2 F_m(X)}{\partial X_i \partial X_j}\right]_{n\times n}, l \leftarrow m$

2: **while** $l = m$ **do**

3: $T(\delta) = F_m(X) + \nabla F_m(X)^T \delta + \dfrac{1}{2}\delta^T \nabla^2 F_m(X)\delta$

    // Use Taylor expansion to approximate $F_m$ in the neighbourhood $U(X, \delta)$ of $X$

4: $L(\delta, \lambda) = F_m(X) + \nabla F_m(X)^T \delta + \dfrac{1}{2}\delta^T \nabla^2 F_m(X)\delta + \lambda(\|\delta\|_p^2 - C)$

    // Construct the Lagrange function

5: $\min -\inf L(\delta, \lambda) \quad s.t.\ \lambda \geq 0$

    // Construct the dual problem

6:     $(\delta^*, \lambda^*) \leftarrow \nabla_\delta L(\delta, \lambda) = 0, \lambda(\|\delta\|_p^2 - C) = 0, \|\delta\|_p^2 - C \leq 0, \lambda \geq 0$

    // The KKT condition is used to find the optimal solution

7:     $l \leftarrow \arg\max_i F_i(X + \delta^*)$

8:     $C \leftarrow C + 0.01$

9: **end while**

10: **return** $X + \delta^*$

---

The selection of $C$ is also involved in generating AEs through equation (18). $C$ is used to constrain $\delta$. If the value of $C$ is too large, the success rate of generating an adversarial example is higher, but the concealment of adversarial example will be weakened, and vice versa. Therefore, the choice of $C$ is crucial. Empirically, the most suitable $C$ is the minimum one which satisfies $\exists t \in I, F_t(X + \delta) > F_y(X + \delta)$ after solving equation (18). We verify this by running our $T_1$ from $C = 0$ on the MNIST and CIFAR10 data sets separately. We plot lines in Figure 2.

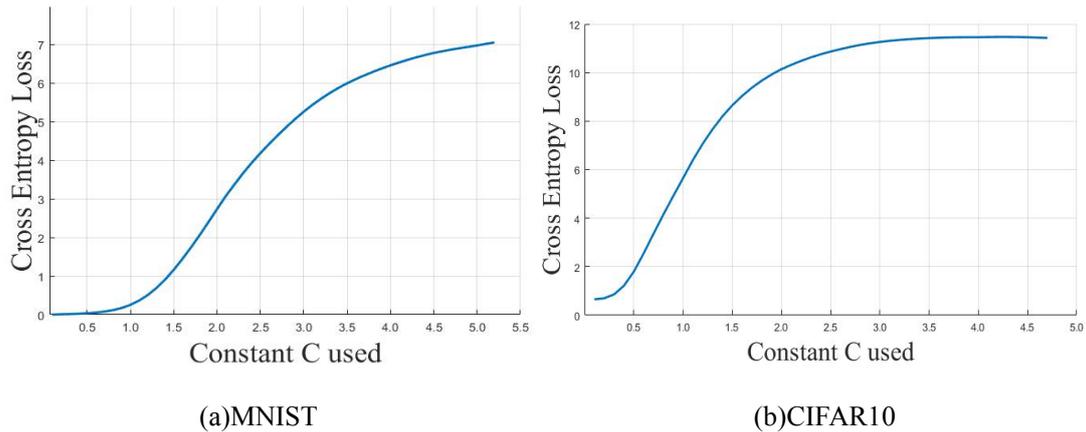

(a)MNIST                    (b)CIFAR10

Figure 2: Change in constant $C$. We plot the relationship between constant $C$ and the change of cross entropy loss value of DNNs, when the objective function $T_1$ generates adversarial examples on MNIST and CIFAR10 dataset respectively.

### 3.3     Generate adversarial examples based on the Gauss-Newton Method

Because of the use of Taylor expansion formula, the method in 3.2 can generate effective and highly transferable adversarial examples, but the shortcomings of this method are also very obvious. Since the Taylor expansion formula involves the calculation of the Hessian matrix, and the second-order term in the Hessian matrix is usually difficult to calculate or requires a large amount of calculation, it is also not advisable to use the secant line approximation of the whole Hessian. Therefore, we can also use gauss-newton method to simplify the calculation.

Gauss-Newton Method is a specialized method for minimizing the least squares cost $(1/2)\|F_y(X + \delta)\|^2$. Given a point $X$, the pure form of the Gauss-Newton Method is based on

linearizing $F_y(X+\delta)$ to obtain

$$R(X+\delta, X+\delta^k) = F_y(X) + \nabla F_y(X)^T(\delta - \delta^k) \tag{21}$$

and then minimizing the norm of the linearized function $R$:

$$\begin{aligned}\delta^{k+1} &= \arg\min_{\delta \in \mathbb{R}^{m\times 1}} (1/2)\|R(X+\delta, X+\delta^k)\|^2 \\ &= \arg\min_{\delta \in \mathbb{R}^{m\times 1}} (1/2)\{\|F_y(X)\|^2 + 2(\delta - \delta^k)^T \nabla F_y(X) F_y(X) \\ &\quad + (\delta - \delta^k)^T \nabla F_y(X) \nabla F_y(X)^T (\delta - \delta^k)\}\end{aligned} \tag{22}$$

Assuming that the matrix $\nabla F_y(X)\nabla F_y(X)^T$ is invertible, the above quadratic minimization yields:

$$\delta^{k+1} = \delta^k - (\nabla F_y(X)\nabla F_y(X)^T)^{-1} \nabla F_y(X) F_y(X) \tag{23}$$

Notice that because of the high nonlinearity of DNNs, we can't prove the matrix $\nabla F_y(X)\nabla F_y(X)^T$ is invertible. To ensure descent, and also to deal with the case where the matrix $\nabla F_y(X)\nabla F_y(X)^T$ is singular(as well as enhance convergence when this matrix is nearly singular), the equation (22) is rewrited as follows:

$$\delta^{k+1} = \delta^k - \alpha^k (\nabla F_y(X)\nabla F_y(X)^T + \Delta^k)^{-1} \nabla F_y(X) F_y(X) \tag{24}$$

where $\alpha^k$ is a stepsize chosen by one of the stepsize rules. The matrix $\nabla F_y(X)\nabla F_y(X)^T$ is a symmetric matrix certainly, so there is a matrix $\Delta^k = -\lambda_{\min}(\nabla F_y(X)\nabla F_y(X)^T)I$ which is a diagonal matrix that makes $\nabla F_y(X)\nabla F_y(X)^T + \Delta^k$ positive definite. The algorithm is as follows:

**Algorithm2** Generate adversarial examples based on the Gauss-Newton Method

$X$ is a benign example

**Input:** $X$, $y_X$, $C$, $\delta = 0$

**Output:** $\delta$

1: **while** $y_{X+\delta} = y_X$ and $\|\delta\|_p \leq C$ **do**

2: $\quad \nabla F_y(X) \leftarrow \left[\dfrac{\partial F_y(X)}{\partial X_i}\right]_{m\times 1}$,

$\quad H \leftarrow \nabla F_y(X) \cdot \nabla F_y(X)^T$,

$\quad \Delta^k \leftarrow -\lambda_{\min}(\nabla F_y(X)\nabla F_y(X)^T)I$

3: $\quad \delta^{k+1} = \delta^k - \alpha^k (H + \Delta^k)^{-1} \nabla F_y(X) F_y(X)$

4: $\quad X \leftarrow X + \delta$

5:   **return** $\delta$

This is an algorithm that can generate adversarial examples quickly, but the accuracy is not high due to the addition of the identity matrix to ensure positive definite. In other words, the concealment of the adversarial examples generated by this algorithm is not as good as we expected. However, we use meaningless images to generate adversarial examples, then we do not have to worry about the problem of concealment. We'll test that experimentally. In addition, 'our method' we mentioned in this paper is the method in Section 3.2.

## 4 Construct robust DNNs through adversarial training

For the purpose of training a robust DNNs reliably, this paper does not use the method that directly focuses on improving the robustness against specific attacks, but first proposes specific requirements that a robust DNNs should be satisfied. Ruitong Huang et al.[12,15,16] described this specific requirement as a min-max optimization problem. On the one hand, we have to find an adversarial version of a given data point $X$ that achieves a high loss. On the other hand, we have to train a model and find the model parameters which minimize the loss of DNNs to adversarial examples. This is exactly the problem to construct robust DNNs through adversarial training.

Szegedy et al.[15] first proposed to use both adversarial examples and benign examples as training data, and the experiment proved that it is an effective method to defense adversarial examples. Aleksander et al.[12], from the perspective of optimization, believed that adversarial training is an optimization problem about saddle points, and they extended traditional ERM training to robust training.

Adversarial training: Suppose $(X, y_{true}) \in D$ is the original training data. Adversarial examples can be obtained under constraint $\varepsilon$ and $J(\cdot)$ is the loss function. Described as follows:

$$\theta^* = \arg\min_{\theta} \mathrm{E}_{(X, y_{true}) \in D} \left[ \max_{\|X^{adv} - X\|_{\infty} \leq \varepsilon} J(f(X^{adv}; \theta), y_{true}) \right] \quad (24)$$

It can be found that (24) is a saddle point problem, a combination of an inner maximization problem and an outer minimization problem. The inner maximization problem is to find adversarial examples that achieves a high loss. The outer minimization problem is to find the model parameters that can minimize the adversarial loss under some kind of adversarial attack. Current work on adversarial examples usually focuses on specific defensive mechanisms, or on attacks against such defenses[12]. An important feature of min-max optimization problem is that attaining small adversarial loss gives a guarantee that targeted attack cannot fool DNNs. By definition, it is possible to construct a robust DNNs which can defense all kinds of attacks. Hence, adversarial training is an optimal balance between model accuracy and robustness.

Equation (24) defines the goal that an ideal DNNs should achieve, and quantifies its robustness. When $\mathbb{E}$ approaching infinity, the corresponding DNNs has perfect robustness against specified attack. This section studies the structure of adversarial training under the background of DNNs. These studies will lead us to use DNNs training to produce DNNs that are highly resistant to a wide range of adversarial attacks. Hence, we now focus our attention on obtaining a good solution to Equation (24).

### 4.1 Inner maximization problem

The inner maximization problem corresponds to constructing valid adversarial examples, which is a non-concave internal maximum problem. Since this problem requires us to maximize a non-concave function, this is difficult to deal with. Our method approximates the output of the non-concave function in the input neighbourhood through the Taylor expansion function, and then we turn this second order unction into a convex optimization problem by using the dual problem. Our method is more conducive to finding the extreme value within the constraint range and avoiding falling into the local optimal solution, which is exactly the defect of the existing typical attack methods.

In order to explain that our method can solve the inner maximization problem effectively, we take MNIST and CIFAR10 dataset as examples, and randomly pick up the pictures that can be correctly classified by DNNs for testing.

Experiment in Section 5 showed that, as we had expected, our method which uses the second-order Taylor expansion function to approximate the output of DNNs, and then uses the dual function to transform not only avoid falling into the local optimal value, and find the global optimal solution within the constraint range, but also make sure that the point $X$ found by the second-order function can also be input into DNNs to get the extreme value.

### 4.2 Outer minimization problem

The previous discussion shows that the inner maximization problem can be solved successfully by using our method. In order to train the adversarial network, we also need to solve the outer minimization problem of equation (6), that is, to find the model parameters to minimize the adversarial loss.

The main method to minimize the loss function is stochastic gradient descent (SGD) when training DNNs. An effective way to calculate the gradient of the outer optimization problem is to calculate the gradient of the loss function at the maximum value of the inner problem. This corresponds to adding the adversarial examples to the original training data set in the adversarial training. Of course, it is not clear that this is a valid descent direction for the min-max optimization problem. However, for continuously differentiable functions ,the Danskin theorem - a classical optimization theorem - states that this is indeed true[12], and that the gradient of the inner maximization problem corresponds to descent directions for the min-max optimization problem. In 5.2, we will prove the effectiveness of the our method for adversarial training through experiments.

# 5 Evaluation

We now use our experimental setup to answer the following questions:(1) Vertical comparison between all objective functions on different dataset;(2) Horizontal comparison between FGSM, JSMA, Deepfool, C&W, PGD, M-DI2-FGSM and our method on different dataset;(3) Can our method improve the robustness of DNNs through adversarial training?(4) Whether Our adversarial examples are sufficiently transferable or not.

## 5.1    Experimental Setup

Our experiment will be conducted on MNIST and CIFAR10 dataset to verify the effectiveness of our method. MNIST is a popular handwritten dataset widely used in the machine learning community. It consists of ten classes from digit 0 to 9, containing a total of 70,000 handwritten digit images. We select 60,000 images as training data and 10,000 images as test data. Each image is in the size of 28×28 pixels.

The CIFAR-10 dataset consists of 60000 32×32 color images in 10 classes, with 6000 images per class. There are 50000 training images and 10000 test images. The dataset is divided into five training batches and one test batch, each with 10000 images. The test batch contains exactly 1000 randomly-selected images from each class.

We use the standard model for each dataset. For MNIST, we use the standard 3-layer convolutional neural network which achieves 99.2% accuracy. For cifar-10, we trained a standard 4-layer convolutional neural network which achieves 95.3% accuracy.

## 5.2    Evaluating Measure

We use $L_p$, PSNR and ASR to measure the effectiveness of our method. The value of $L_p$ is generally used to measure the global or local added perturbation, which is a measure of the concealment of adversarial examples. In order to better evaluate the concealment of adversarial examples, we listed "Peak Signal to Noise Ratio" (PSNR) as one of the indicators. PSNR, as the most common and widely used objective measurement of image quality, can effectively evaluate the concealment of adversarial examples. ASR stands for the probability of success in generating adversarial examples. When ASR is not 100%, $L_p$ and PSNR are for successful attacks only.

A. Vertical comparison between objective functions

|       | MNIST | | | CIFAR10 | | |
|-------|-------|------|------|------|-------|------|
|       | $L_2$ | PSNR | ASR  | $L_2$ | PSNR | ASR  |
| $T_1$ | 0.91  | 73.03| 100% | 0.29 | 87.41 | 100% |
| $T_4$ | 1.40  | 71.46| 100% | 0.30 | 87.45 | 100% |

Table 1. Evaluation of untargeted attack by different objective functions on MNIST and CIFAR10 dataset. We show the average $L_2$ distortion, $PSNR$ and $ASR$ of the objective function $T_1$ and $T_4$.

|         |       | MNIST | | | CIFAR10 | | |
|---------|-------|-------|-------|-------|-------|-------|-------|
|         |       | $L_2$ | PSNR  | ASR   | $L_2$ | PSNR  | ASR   |
|         | $T_2$ | 1.24  | 71.60 | 100%  | 0.39  | 84.83 | 100%  |
| Best    | $T_3$ | 1.74  | 70.27 | 100%  | 0.41  | 71.70 | 100%  |
| Case    | $T_5$ | 0.99  | 71.41 | 100%  | 0.54  | 82.16 | 100%  |
|         | $T_6$ | 1.24  | 70.45 | 100%  | 0.42  | 84.75 | 100%  |
|         | $T_2$ | 1.69  | 68.38 | 100%  | 0.66  | 82.86 | 100%  |
| Average | $T_3$ | 1.74  | 69.10 | 75.2% | 0.39  | 81.14 | 70.1% |
| Case    | $T_5$ | 1.70  | 67.21 | 100%  | 0.54  | 81.30 | 100%  |
|         | $T_6$ | 1.61  | 67.98 | 78.5% | 0.62  | 80.12 | 73.5% |
|         | $T_2$ | 2.49  | 65.95 | 100%  | 0.54  | 82.13 | 100%  |
| Worst   | $T_3$ | 2.50  | 67.43 | 74.5% | 0.57  | 80.08 | 69.7% |
| Case    | $T_5$ | 2.24  | 63.75 | 100%  | 0.56  | 80.52 | 100%  |
|         | $T_6$ | 2.25  | 65.04 | 75.6% | 0.61  | 82.14 | 71.2% |

Table 2. Evaluation of targeted attack by different objective functions on MNIST and CIFAR10 dataset. We show the average $L_2$ distortion, $PSNR$, and $ASR$ of the objective function $T_2$, $T_3$, $T_5$ and $T_6$.

Table 1 and Table 2 show the experimental results. In Table 2, Best Case, Worst Case and Average Case represent performing the attack against all incorrect classes, and then report the target class that was least difficult to attack, most difficult to attack and a random one among the labels that are not the correct label, respectively. We evaluated the quality and success rate of the adversarial examples generated by six objective functions on MNIST and CIFAR10 dataset. In untargeted attack, the only difference between $T_1$ and $T_4$ is that the gradient information of $T_1$ comes from a hidden layer at the end, while the gradient information of $T_4$ comes from the Softmax layer. In targeted attack, the difference between $T_2$, $T_3$ and $T_5$, $T_6$ is the same. Experimental results show that the objective function locates at the last hidden layer and locates at the Softmax layer doesn't make very much difference. Under the influence of the normalization of Softmax layer, $T_4$, $T_5$ and $T_6$ perform even better than $T_1$, $T_2$ and $T_3$ under the same conditions. It suggests that in our method, the normalization caused by the Softmax layer[9] does not reduce the quality of information about how neurons are activated by different inputs. Therefore, whether or not the defense method smooths the gradient of DNNs[21], we can get adversarial examples.

B. Horizontal comparison between existing classic methods and our method

To verify the effectiveness of our adversarial examples, we used JSMA, C&W, FGSM, Deepfool and M-DI$^2$-FGSM for comparison, where codes of JSMA, C&W, FGSM and Deepfool come from Cleverhans[22] and the codes for PGD and M-DI$^2$-FGSM come from the link given by the author in the original text [12,13]. In addition, to ensure the rigor of the evaluation, we use the same model and the same batch of test data to verify the above methods.

For FGSM, we take $\varepsilon = 0.01$. If the target class adversarial examples can be generated within the specified step size, the adversarial examples will be returned for evaluation, otherwise it will be regarded as a failure. As a derivative method of FGSM, PGD has an upper limit $\varepsilon = 8.0$ for each pixel on the pixel scale of "0-255". For JSMA, we aim to generate adversarial examples. We extend the constraint on perturbation, and modify the iteration termination condition to classify as the target class successfully. That is, no matter how much perturbation is required, we report success if the attack produce adversarial examples with the correct target label. But JSMA is unable to run on CIFAR10 due to an inherent significant computational cost for searching saliency map[11]. If we remove the search process, JSMA's ability to generate adversarial examples is greatly reduced. Therefore, we did not use JSMA in the CIFAR10 experiment. Note that CW is a bit different from the above gradient-based methods in that it is an optimization-based attack. In this experiment, $L_2$ norm attack in C&W is adopted, and we set $\varepsilon = 1$, learning rate = 0.1. For Deepfool, note that in our implementation, the noise calculated as $f/\|w\| * w$ instead of $f/\|w\| * w/\|w\|$, where $\|w\|$ is the $L_2$ norm.

In our experiment, 500 pictures that could be correctly judged by the initial model were randomly selected from MNIST and CIFAR10 for testing. After all, if an image can be misclassified without a perturbation, then the meaning of generate adversarial examples is lost. In addition, in the case of target attack, we also divided adversarial examples into the best case and the worst case according to the image quality of the adversarial examples. In other words, it is to compare the perturbation strength superimposed by different methods to generate adversarial examples. The results are shown in the following table.

|  |  | MNIST |  |  | CIFAR10 |  |  |
|---|---|---|---|---|---|---|---|
|  |  | $L$ | PSNR | ASR | $L$ | PSNR | ASR |
| Best Case | Our $L_0$ | 239.1 | 71.21 | 100% | - | - | - |
|  | JSMA | 183.5 | 58.54 | 100% | - | - | - |
|  | Our $L_2$ | 1.70 | 67.21 | 100% | 0.39 | 84.83 | 100% |
|  | C&W | 1.04 | 61.66 | 100% | 0.28 | 63.26 | 100% |
|  | Our $L_\infty$ | 1.34 | 70.86 | 100% | 1.32 | 70.56 | 100% |
|  | FGSM | 4.12 | 74.47 | 62.2% | 1.72 | 64.72 | 100% |
| Average Case | Our $L_0$ | 243.4 | 74.04 | 100% | - | - | - |
|  | JSMA | 199.6 | 58.66 | 100% | - | - | - |
|  | Our $L_2$ | 1.84 | 67.21 | 100% | 0.50 | 82.86 | 100% |
|  | C&W | 2.21 | 60.68 | 96.2% | 0.62 | 65.52 | 100% |
|  | Our $L_\infty$ | 2.32 | 70.90 | 100% | 2.10 | 69.74 | 100% |
|  | FGSM | 5.09 | 73.10 | 45.3% | 1.78 | 63.75 | 89.5% |
| Worst Case | Our $L_0$ | 332.4 | 72.13 | 100% | - | - | - |
|  | JSMA | 265.4 | 57.67 | 100% | - | - | - |
|  | Our $L_2$ | 2.24 | 69.75 | 100% | 0.54 | 82.13 | 100% |
|  | C&W | 3.30 | 58.80 | 100% | 0.35 | 59.54 | 100% |
|  | Our $L_\infty$ | 2.20 | 61.34 | 100% | 3.2 | 65.12 | 100% |
|  | FGSM | 5.21 | 65.24 | 38.2% | 1.92 | 61.80 | 76.3% |

Table 3. Comparison of three targeted attack algorithms on the MNIST and CIFAR10 dataset.

|  | MNIST |  |  | CIFAR10 |  |  |
|---|---|---|---|---|---|---|
|  | $L_2$ | PSNR | ASR | $L_2$ | PSNR | ASR |
| $T_1$ | 0.91 | 73.03 | 100% | 0.29 | 87.41 | 100% |
| $T_4$ | 1.40 | 71.46 | 100% | 0.30 | 87.45 | 100% |
| PGD | 5.17 | 67.72 | 100% | 1.63 | 78.76 | 86.49% |
| Deepfool | 1.66 | 73.80 | 88.12% | 0.16 | 85.20 | 81.44% |
| M-DI²-FGSM | 3.14 | 65.12 | 56.6% | 1.85 | 75.12 | 48.1% |

Table 4. Comparison of three untargeted attack algorithms on the MNIST and CIFAR10 dataset.

We use the $L_p$ norm and PSNR value to measure the concealment after adding perturbation. Experiments show that on different dataset, compared with the existing classical attack methods, the adversarial examples generated by our method have better imperceptibility, and our method can produce target class adversarial examples for any picture.

In [9], JSMA uses the last hidden layer instead of the Softmax layer to calculate the adversarial saliency map. The essence of this approach is to iteratively modify the pixels with the maximum derivative value until adversarial examples is generated or the number of pixels modified exceeds the limit. The authors gave a simple example to show how small input perturbations found using the forward derivative can induce large variations of the neural network output, but they didn't explain the mathematical derivation. We believe that the mathematical basis of this method comes from the fact that in the neighbourhood of a fixed value $X$, DNNs satisfy: For small $\|\delta\|$, there is $F(X+\delta) \simeq \ell(\delta) \equiv F(X) + \nabla F(X) \cdot \delta$. Therefore, JSMA can generate adversarial examples by searching the adversarial saliency map.

However, authors believe that the extreme variations introduced by the Softmax layer lead to extreme derivative values. This reduces the quality of information on how the neurons are activated by different inputs and causes the forward derivative to loose accuracy when generating saliency maps. Therefore they compute the forward derivative of the network using the last hidden layer instead of the Softmax layer. According to equation (3), the author selected the pixel with the maximum value of $S(X,t)$, but essentially wanted to find the pixel with the maximum value of $\partial F_t(X)/\partial X_i$ as the pixel most conducive to classification as the target class $t$ after adding perturbation. However, without the normalization of the Softmax layer, the author cannot guarantee that the increase of $\partial F_t(X)/\partial X_i$ will bring the decrease of $\left|\sum_{j \neq i} \partial F_t(X)/\partial X_i\right|$. So the greater the value of $S(X,t)$ doesn't mean the greater the value of $\partial F_t(X)/\partial X_i$. This means that the pixels selected in the above equation are not technically the key pixels in the targeted attack. We have reason to believe that this is why the capability of JSMA is not as good as we expected. However, our method overcomes this disadvantage. Whether it is in the last hidden layer or in the Softmax layer, the our method can effectively modify pixel points and effectively generating adversarial examples.

In [11], C&W is committed to solving $\text{minimize}\|\delta\|_p + c \cdot f(X+\delta) \ s.t. X+\delta \in [0,1]^n$. The authors use binary search to determine the value of constant $c$, which is a mechanical search method that is far less accurate and flexible than the Lagrange multiplier method we use. At the same time, C&W uses $\delta_i = \frac{1}{2}(\tanh(w_i)+1) - x_i$ to expand the search space. Admittedly, this method is very conducive to searching more powerful adversarial examples, but it will also bring a very large search cost. While our method can be successful without having to pay such a high price.

In [12], PGD transforms constrained optimization problem into unconstrained optimization problem, which is easy to implement and suitable for solving large-scale optimization problems to generate effective adversarial examples. However, in order to ensure the effectiveness of iteration, it takes a long time to calculate the projection of iteration points and the convergence rate is slow. In addition, when PGD iterates the iteration point from outside to inside of the constraint through Equation (8), part of the iteration information is inevitably lost. In our paper, after approximating the input output mapping relationship of DNNs by using quadratic Taylor expansion, the dual function is constructed by using Lagrange multiplier method to optimize and generate adversarial examples. The Lagrange multiplier method can also transform unconstrained optimization problems into constrained optimization problems. However, unlike the projected gradient descent algorithm, the Lagrange multiplier method does not lose iteration information due to iteration, so the optimization results are more accurate.

Both Deepfool and FGSM often get trapped in local optimum due to their algorithmic characteristics, so the global optimal solution cannot be obtained. In this paper, quadratic Taylor expansion is used to approximate the input output mapping relationship of DNNs, which can effectively skip the local optimum. M-DI$^2$-FGSM combines momentum and diverse inputs naturally to form an attack with stronger transferability, but experiments have proved that its ASR is not ideal.

C. Generate synthetic digits

Based on the above experiments, we found that we can make any picture into adversarial examples of target class, this theory is also applicable to meaningless pictures. [9] and [11] have both done such experiments. They use all-black image and all-white image to generate adversarial examples that make no sense to humans but misclassified by DNNs . Here is the result.

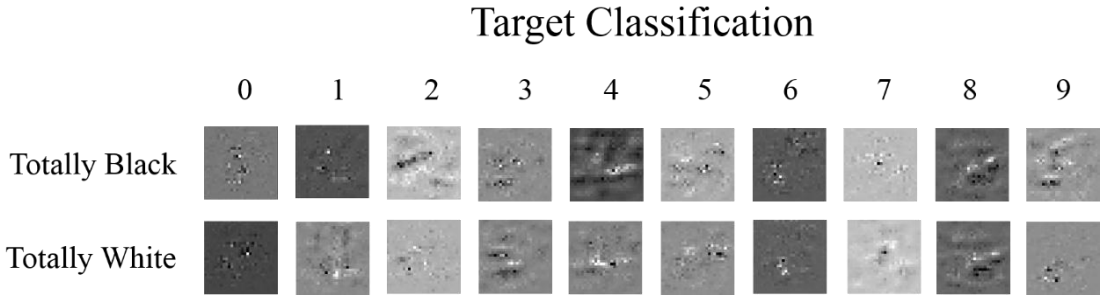

Figure 5: Targeted attack for the MNIST where the starting image is totally black or white. For random synthetic digits in [9] ,one can clearly recognize the target digit, but method in Section 3.3 does not have this flaw. For random synthetic digits in [11], although the perturbation is very small, its calculation cost is very large, which is greatly inferior to the method in Section 3.3.

## 6 Related work

Since Szeigy et al.[7] discovered adversarial examples, the safety of DNNs[24] has become an active research topic. The researchers classified the attack and discussed the adversarial capabilities[25,26]. Szegedy et al.[7] proposed a box-constrained LBFGS method to generate adversarial examples. Goodfellow et al.[14] proposed FGSM to generate adversarial examples efficiently by performing a single gradient step. Kurakin et al.[23] extended it to an iterative version, and found that the adversarial examples also exist in the physical world. Dong et al.[27] proposed a broad class of momentum-based iterative algorithms to boost the transferability of adversarial examples.

The above work calculates the gradient of DNNs to generate adversarial examples[7,9,14,28,29]. These work calculates the gradient not to update the weight of DNNs to improve the network, but to update the input and then make DNNs misclassify. They first define the cost function for the output of DNNs and then optimize the cost function to generate adversarial examples. However, these cost functions are often difficult to calculate or have to bear a large computational cost to optimize. Unlike these methods, our method uses a quadratic Taylor expansion to approximate the complicated input output mapping relationship of DNNs in a small neighbourhood of the input example directly. This is an equation that can directly find the extremum in the neighborhood, which is more accurate and requires less computational cost. Therefore, our method can generate adversarial examples which are very effective under both white-box and black-box settings.

The existence of adversarial examples reveals the vulnerability of DNNs. In order to improve the robustness of DNNs, researchers have proposed many methods of square defense adversarial examples.[14,30] proposed to use both adversarial examples and benign examples as training data to increase the robustness of DNNs. Tram`er et al.[31] proposed ensemble adversarial training, which augments training data with perturbations transferred from other models, in order to improve the robustness of DNNs. We recommend using counter training as a defense. Adversarial training requires a large number of effective adversarial examples to be generated at low cost, which is the main purpose of this paper. Min-max optimization problem is considered in confrontational training[32,33]. However, the results mentioned in [32,30] are different from those in our paper. Firstly, the authors believes that the inner maximization problem is difficult to solve, and the innovation of the method in this paper overcomes this problem. We approximate the output of DNNs, then the Lagrange multiplier method is used to construct the dual function for optimization. It is proved theoretically and experimentally that our method can obtain the optimal solution for the inner maximization problem. Secondly, they only considered first-order adversarial, and we also experimented with multi-step iterative methods. Furthermore, although the experiments in [33] produced promising results, they were evaluated only on the basis of FGSM. However, the assessment limited to FGSM is not entirely reliable. Therefore, our method is compared with many methods to obtain a more reliable experimental result.

# 7  Conclusion

This paper proposes a novel and more powerful attack method. We use a quadratic Taylor expansion to approximate the input output mapping relationship of DNNs in a small neighbourhood of the input example (using $L_p$ norms constraints) , replacing the nonlinear part of DNNs. After that, the Lagrange multiplier method is used to construct the dual function for optimization and calculate the extreme value to generate adversarial examples. This method can efficiently generate a large number of effective adversarial examples at a small cost to solve the inner maximization problem in the min-max optimization problem.

Experimental results on MNIST and CIFAR10 show that compared with the existing classical attack methods, our method is more covert, more transferable, and can significantly improve the robustness of DNNs through confrontation training. Compared with single-step attack, our method has high transferability while maintaining the concealment of the sample. Compared with iterative attack, our method can solve the internal maximization problem more effectively and accurately. Therefore, our proposed method can be used as a benchmark to evaluate the robustness of DNNs against opponents and the effectiveness of different defense methods in the future.

In the future work, we will improve this attack method and extend it to cyclic recursive neural network instead of the periodic neural network considered in this paper. We will continue to study the root cause of the existence of the adversarial example, hoping to propose a more robust DNNs training model.